\documentclass[10pt, a4paper]{article}

\usepackage{lrec-coling2024} 

\usepackage{verbatim}
\usepackage{graphicx}  
\usepackage{multirow}
\usepackage{enumerate}
\usepackage{float}
\usepackage{amsmath}
\usepackage{color}
\usepackage{booktabs}
\usepackage{url}
\usepackage{enumitem}
\usepackage{appendix}

\title{Exploring the Necessity of Visual Modality \\
in Multimodal Machine Translation using Authentic Datasets}

\name{Zi Long$^1$, Zhenhao Tang$^2$, Xianghua Fu$^1$, Jian Chen$^2$, Shilong Hou$^2$, Jinze Lyu$^2$} 

\address{$^1$College of Big Data and Internet, Shenzhen Technology University, Shenzhen, China \\
$^2$College of Application and Technology, Shenzhen University, Shenzhen, China\\}

\abstract{
Recent research in the field of multimodal machine translation (MMT) has indicated that the visual modality is either dispensable or offers only marginal advantages.
However, most of these conclusions are drawn from the analysis of experimental results based on a limited set of bilingual sentence-image pairs, such as Multi30k.
In these kinds of datasets, the content of one bilingual parallel sentence pair must be well represented by a manually annotated image, which is different from the real-world translation scenario. 
In this work, we adhere to the universal multimodal machine translation framework proposed by~\citet{tang2022_multimodal}.
This approach allows us to delve into the impact of the visual modality on translation efficacy by leveraging real-world translation datasets.
Through a comprehensive exploration via probing tasks, we find that the visual modality proves advantageous for the majority of authentic translation datasets.
Notably, the translation performance primarily hinges on the alignment and coherence between textual and visual contents.
Furthermore, our results suggest that visual information serves a supplementary role in multimodal translation and can be substituted.
 \\ \newline  }

\begin{document}
\maketitleabstract


\section{Introduction}
\label{sec:intro}
With the development of neural machine translation (NMT), the role of visual information in machine translation has attracted researchers' attention~\citep{specia2016shared,elliott2017findings2,barrault2018findings3}.
Different from those text-only NMT~\citep{bahdanau2014attn,gehring2016convolutional}, a bilingual parallel corpora with manual image annotations are used to train an MMT model by an end-to-end framework, and therefore visual information can assist NMT model to achieve better translation performance~\citep{calixto2017incorporating,calixto2017doubly,su2021bi-co}.

Concurrently, researchers have also undertaken a diverse range of experiments in an effort to validate the specific role of visual information in NMT.
For example,~\citet{gronroos-etal-2018-memad} and~\citet{lala-etal-2018-sheffield} observed that the robustness of MMT systems remains unaffected when the input image lacks direct relevance to the accompanying text. Notably, the absence of visual features, as highlighted by~\citet{elliott2018adversarial}, also does not yield detrimental effects.~\citet{wu-etal-2021-good} underscores that the utilization of the visual modality serves as a regularization mechanism during training rather than serving as a true complement to the textual modality.
Oppositely,~\citet{caglayan2019probing} delve into the correlation between visual features and text. Their investigation reveals that incorporating the input image aids translation, particularly when certain input words are masked.~\citet{li-etal-2022-vision} design more detailed probing tasks and found that stronger vision features strengthen MMT systems.

Note that most of the previous conclusions are drawn from the analysis of experimental results based on a restricted selection of manually annotated bilingual sentence-image pairs, known as the Multi30k dataset~\citep{elliott2016Multi30K}.
Within the Multi30k dataset, as depicted in Table~\ref{tab:dataset}, the sentences primarily comprise common and straightforward vocabulary, with each bilingual parallel sentence pair being effectively depicted by a single image.   
Table \ref{tab:dataset} also presents an illustration of a bilingual sentence-image pair extracted from a genuine news report from the United Nations News\footnote{
  \url{https://news.un.org/en/}
}, alongside examples of sentence pairs derived from other other authentic translation datasets. Evidently, a substantial disparity exists between the Multi30k dataset and the authentic translation data.
Hence, the evidence and findings derived from Multi30k may potentially exhibit inadequate generalizability and offer limited utility when attempting to analyze the role of the visual modality in MMT within real-world translation scenarios.
In these scenarios, sentences often incorporate rare and uncommon words and are only partially depicted by accompanying images. 

\begin{table*}[t]
    \centering
    \begin{tabular}{|c||c|c|}
        \hline
        Data source & Sentences & Image  \\ \hline \hline
        Multi30k 
        & 
         \begin{tabular}{cp{6cm}}
             \!\!\!EN: \!\!\!& A dog is running in the snow. \\
             \!\!\!DE: \!\!\!& Ein Hund rennt im Schnee.
         \end{tabular}
        & 
        \!\!\!
        \begin{tabular}{c}
            \specialrule{0em}{1pt}{1pt}
            \!\!\!
            \includegraphics[scale=0.6]{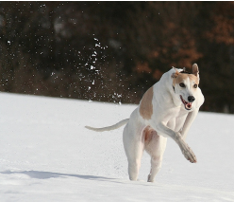}
            \!\!\!
        \end{tabular} 
        \!\!\!
         \\ \hline
         UN News 
        & 
         \begin{tabular}{cp{6cm}}
             \!\!\!EN: \!\!\!& Rescue workers look for survivors in a building in Samada, Syria destroyed by the February 6 earthquake. \\
             \!\!\!DE: \!\!\!& Rettungskr\"{a}fte suchen nach \"{U}berlebenden in einem Geb\"{a}ude in Samada, Syrien, das durch das Erdbeben vom 6. Februar zerst\"{o}rt wurde.
         \end{tabular}
        & 
        \!\!\!
        \begin{tabular}{c}
            \specialrule{0em}{1pt}{1pt}
            \!\!\!
            \includegraphics[scale=0.43]{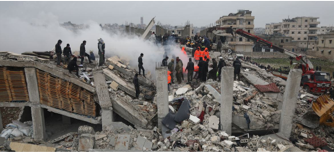}
            \!\!\!
        \end{tabular} 
        \!\!\!
         \\ \hline
         Bible 
         &
          \begin{tabular}{cp{6cm}}
             \!\!\!EN: \!\!\!& I saw, and behold, there was no man,and all the birds of the sky had fled. \\
             \!\!\!DE: \!\!\!& Ich sah, und siehe, da war kein Mensch, und alle V\"{o}gel unter dem Himmel waren weggeflogen.
         \end{tabular}
         & 
         no image 
         \\ \hline
         MultiUN 
         &
          \begin{tabular}{cp{6cm}}
             \!\!\!EN: \!\!\!& Development assistance cannot by itself prevent or end conflict. \\
             \!\!\!DE: \!\!\!& Entwicklungshilfe allein kann Konflikte weder verh\"{u}ten noch beenden.
         \end{tabular}
         & 
         no image 
         \\ \hline
    \end{tabular}
    \caption{Comparison between Multi30k Dataset and Authentic Datasets}
    \label{tab:dataset}
\end{table*}

In a recent study,~\citet{tang2022_multimodal} introduced a universal multimodal neural machine translation model that integrates open-vocabulary image retrieval techniques. 
%
In this work, inspired by ~\citet{tang2022_multimodal}, 
we formulate a set of comprehensive probing tasks aimed at assessing the extent to which the visual modality enhances MMT within real-world translation scenarios.
In addition to commonly used Multi30k, we conduct an extensive set of experiments across four authentic text-only translation datasets. 
We further evaluated two visual noise filtering approaches based on the correlation between textual and visual content.
Furthermore, we investigate the necessity of visual modality in the current multimodal translation process by substituting visual data with closely equivalent textual content.
To summarize, our findings are:
\begin{description}
\item[(1)] Visual modality is mostly beneficial for translation, but its effectiveness wanes as text vocabulary becomes less image-friendly.

\item[(2)] The MMT performance depends on the consistency between textual and visual contents, and utilizing filters based on the textual-visual correlation can enhance the performance.

\item[(3)] Visual information plays a supplementary role in the multimodal translation process and can be substituted by the incorporation of additional textual information.
\end{description}

\section{Related Work}
\label{sec:rel}

The integration of extra knowledge to build fine-grained representations is a crucial aspect in language modeling~\citep{li2020explicit,li2020data,zhang2020semantics}. Incorporating the visual modality into language modeling has the potential to enhance the machine's understanding of the real world from a more comprehensive perspective. 
Inspired by the studies on the image description generation task~\citep{elliott2015multilingual,venugopalan2015sequence,xu2015show}, MMT models have gradually become a hot topic in machine translation research. 
In some cases, visual features are directly used as supplementary information to the text presentation. For example, \citet{huang2016attention} take global visual features and local visual features as additional information for sentences. \citet{calixto2017incorporating} initializes the encoder hidden states or decoder hidden states through global visual features.
\citet{calixto2017doubly} use an independent attention mechanism to capture visual representations. \citet{caglayan2016multimodal} incorporate spatial visual features into the MMT model via an independent attention mechanism. On this basis, \citet{delbrouck2017compact} employs compact bilinear pooling to fuse two modalities. \citet{lin2020dynamic} attempt to introduce the capsule network into MMT, they use the timestep-specific source-side context vector to guide the routing procedure. \citet{su2021bi-co} introduce image-text mutual interactions to refine their semantic representations.

\begin{figure*}[th]
    \centering
    \includegraphics[scale=0.49]{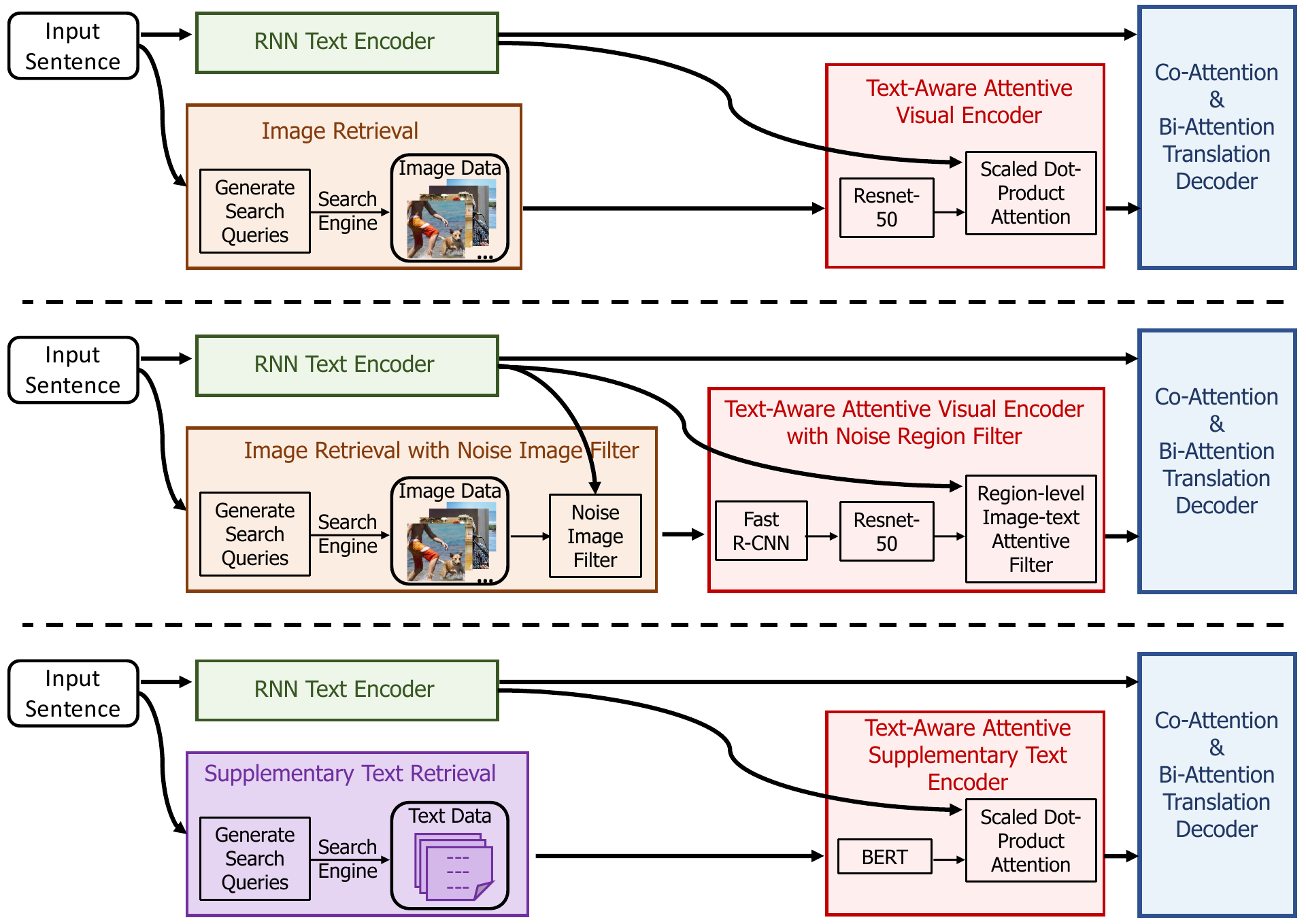}
    \caption{Frameworks of three probing methods}
    \label{fig:flow_3_in_1}
\end{figure*}

Researchers have also come to recognize the potential redundancy of the visual modality. Inconsequential images exhibit minimal impact on translation quality, and the absence of images does not yield a significant drop in BLEU scores, as noted by~\citet{elliott2018adversarial}. Encouraging findings emerged in the study by~\citet{caglayan2019probing}. They highlighted the continuing utility of the visual modality in scenarios where linguistic context is limited but noted its diminished sensitivity when exposed to complete sentences.
In a more recent investigation,~\citet{wu-etal-2021-good} attributed the observed BLEU improvement in MMT tasks to training regularization. They underscored the importance of constructing appropriate probing tasks with inadequate textual input. It's important to highlight that the proposed probing task represents an enhanced iteration building upon prior research~\citep{caglayan2019probing, wu-etal-2021-good}.
~\citet{li-etal-2022-vision} made a systematic study on whether stronger vision features are helpful.
All the preceding research has been conducted exclusively on the Multi30k dataset, which has limitations in scale and considerably differs from real-world translation scenarios.
In this study, we employ the framework introduced by ~\citet{tang2022_multimodal} to systematically examine the influence of visual information across various authentic translation datasets, extending our analysis beyond the limitations of the small and specialized Multi30k dataset.

\section{Preliminary}
\label{sec:models}


We start with a description of three probing methods employed in this work, which encompass the approach introduced by ~\citet{tang2022_multimodal} and two additional methods derived from it. Figure ~\ref{fig:flow_3_in_1} shows frameworks of these three methods.

\subsection{MMT with Search Engine Based Image Retrieval}
\label{sec:tang_mmt}

As depicted in the top section of Figure ~\ref{fig:flow_3_in_1}, ~\citet{tang2022_multimodal} introduced a search engine-based image retrieval technique and a text-aware attention image encoder. This innovation enables the handling of authentic text-only translation data within MMT systems.
We implement this approach across multiple authentic translation datasets to examine the influence of visual information across datasets with varying styles.
To ensure the comprehensiveness of this paper, this section will provide a brief overview of the approach proposed by Tang et al. (2022).

\paragraph{Text Encoder}
\label{sec:text_enc}

In this work, we employ a commonly utilized bi-directional LSTM as the RNN text encoder.
For a given sentence denoted as $X$, the output of the text encoder is represented as $C = (\boldsymbol{h}_{1}, \boldsymbol{h}_{2}, \ldots, \boldsymbol{h}_{N})$, where $N$ denotes the length of the sentence $X$.

\paragraph{Image Retrieval}
\label{sec:img_ret}

To emphasize the core components of the sentence and mitigate the impact of noise, including stopwords and infrequent words, ~\citet{tang2022_multimodal} utilized the TF-IDF method ~\citep{witten2005kea} to generate search queries for image search engines.
Subsequently, the generated search queries are utilized in image search engines to retrieve the first available image associated with each query.
For each given sentence $X$, $M$ search queries denoted as ($q_{1}, q_{2}, \ldots, q_{M}$) are generated, and subsequently $M$ images represented as ($A_{1}, A_{2}, \ldots, A_{M}$) are retrieved from search engines.

\paragraph{Text-Aware Attentive Visual Encoder}
\label{sec:img_enc}
 
Each image $A_m$ ($m = 1, \dots, M$) is transformed into a $196 \times 1024$ dimensional feature vector using ResNet-50~\citep{he2016res50} .
A simple but effective scaled dot-product attention in visual encoder is subsequently employed in the visual encoder to derive a resultant visual representation. Here, we utilize the average pooling $C_{'}$ of the text representation $C = (\boldsymbol{h}_{1},\boldsymbol{h}_{2},\ldots,\boldsymbol{h}_{N})$ as the query, while the visual feature vectors $A_{1}, A_{2}, \ldots, A_{M}$ serve as the keys and values in this attention mechanism.
The resultant visual representation $A$ is also expressed as a $196 \times 1024$ dimensional feature vector, which can be regarded as a matrix $A = (\boldsymbol{a}_{1}, \boldsymbol{a}_{2},\ldots,\boldsymbol{a}_{L})$, where $L=196$ and each $\boldsymbol{a}_{l}\in R^{1024}$ ($l=1,\dots,L$). 
Visual representation $A = (\boldsymbol{a}_{1}, \boldsymbol{a}_{2},\ldots,\boldsymbol{a}_{L})$ and text representation $C = (\boldsymbol{h}_{1},\boldsymbol{h}_{2},\ldots,\boldsymbol{h}_{N})$ are then used as the inputs of translation decoder.

\paragraph{Translation Decoder}
\label{sec:dec}

For the decoder, we adopt the approach introduced by ~\citet{su2021bi-co}, implementing both a bidirectional attention network and a co-attention network to effectively capture the underlying semantic interactions between textual and visual elements.
 Based on the results of the preliminary experiment, it was evident that transformer-based models did not confer a performance advantage on datasets like Global Voices and other smaller ones. Consequently, we followed the approach of~\citet{tang2022_multimodal} and selected LSTM as our foundational model.
%
The bidirectional attention network enhances the representations of both text and image. 
These enhanced representations are subsequently input into the co-attention network to obtain the time-dependent context vector $c_{t}$ and the visual vector $v_{t}$. 
Within the co-attention network, we calculate the probability distribution for the next target word $y_{t}$ using the previous hidden state $s_{t-1}$, the previously generated target word $y_{t-1}$, the time-dependent context vector $c_{t}$, and the time-dependent visual vector $v_{t}$.

\subsection{MMT with Visual Noise Filtering}
\label{sec:filter_method}

Considering that the noise images obtained from search engines could have a substantial impact on the performance of the MMT system, we further evaluated two visual noise filtering approaches based on the correlation between textual and visual content, as depicted in the central part of Figure~\ref{fig:flow_3_in_1}. 
One approach utilizes the pretrained CLIP model to filter out noise images, while the other employs a region-level image-text attentive filter module to filter out noisy image regions.

\paragraph{Noise Image Filter}
\label{sec:img_filter}

In the CLIP-based noise image filtering approach, we begin by retrieving $M^{'}$ ($M^{'} > M$) images from search engines for each input sentence. 
Following this, we calculate the correlation between the input text and the retrieved images using a pretrained CLIP model ~\citep{radford2021learning}. 
Subsequently, we select only the top-$M$ images with the highest correlation to the input source text as the output of the image retrieval process.

\paragraph{Noise Region Filter}
\label{sec:region_filter}

In the noise image region filtering approach, we begin by extracting convolutional feature maps from the top-$O$ most confident regions denoted as ($r_{1}, \dots, r_{O}$) in each collected image.
This is achieved using a pretrained Faster R-CNN model~\citep{ren2015faster}, aiding in the initial filtration of visual information that may be challenging to distinguish as distinct regions in the images.
The image region of each collected image is then represented as a $1024$ dimensional feature vector using ResNet-50. For all the retrieved $M$ images, we extract a total of $M \times O$ regions ($r_{1}, \dots, r_{M \times O}$), resulting in $M \times O$ feature vectors ($\boldsymbol{a}_{1}, \dots, \boldsymbol{a}_{M \times O}, \boldsymbol{a}_{o} \in R^{1024}$).
Subsequently, we compute the correlation score between each image region and the input text using the following equation:
\begin{eqnarray}
 S(\boldsymbol{a}_{o}, C^{'}) & = & V_{a} {\rm tanh} (W_{a} \boldsymbol{a}_{o} + U_{a} C^{'} ) \nonumber
\end{eqnarray}
Here, $C^{'}$ represents the average pooling of the text representation $C = (\boldsymbol{h}_{1}, \boldsymbol{h}_{2},\ldots,\boldsymbol{h}_{N})$.
We retain only the visual information from the top-$O$ most relevant regions out of the initially extracted $M \times O$ regions. 
This preserved visual information serves as the visual representation for the given input sentence, denoted as
$A = \{ \boldsymbol{a}_{o} | {S}(\boldsymbol{a}_{o}, C^{'})$ ranks in the top-$O, 1 \leq o \leq M \times O\}$
, and it is subsequently fed into the translation decoder module. Less relevant regions are discarded during this process.

\subsection{Supplementary Text Enhanced NMT}
\label{sec:text_support}

As discussed by ~\citet{caglayan2019probing}, multimodal translation models typically view visual information as a complementary component to textual information. 
However, we raise the question of whether this complementary role can also be achieved by incorporating additional textual information, potentially obviating the need for images in the process.
Hence, our investigation aims to assess the necessity of visual information in the existing multimodal translation process by substituting visual data with nearly equivalent textual information. 
As illustrated in the lower section of Figure~\ref{fig:flow_3_in_1}, we replace the image retrieval module with a supplementary text retrieval module and substitute the text-aware attentive visual encoder with a similar text-aware attentive supplementary text encoder.

\paragraph{Supplementary Text Retrieval}
\label{sec:text_ret}

Similar to the process of retrieving images from search engines, we collected supplementary textual data from search engines. 
For every input source sentence $X$, we follow the same approach as outlined in Section~\ref{sec:img_ret} to generate $M$ search queries ($q_{1}, \dots, q_{M}$). 
Subsequently, we collect $M$ sentences ($T_{1}, \dots, T_{M}$) that contains all the terms present in the respective search queries ($q_{i} \subseteq T_{i}, 1 \leq i \leq M$). 

\paragraph{Text-Aware Attentive Supplementary Text Encoder}
\label{sec:supp_text_enc}

Each gathered supplementary text $T_m$ ($m=1, \dots, M$) is transformed into a $N \times 1024$ dimensional feature vector using BERT~\citep{devlin2018bert}, where $N$ denotes the length of the gathered text data.
To ensure consistency, these textual feature vectors are subsequently padded to match the dimensions of $L \times 1024$ ($L=196$), aligning them with the visual feature vectors. 
These feature vectors are then integrated into the scaled dot-product attention module as keys and values, with the average pooling $C^{'}$ representing the input text serving as the query. 
The resultant supplementary text representation is then passed to the translation decoder.

\section{Experiment Setup}
\label{sec:exper}

\subsection{Dataset}
\label{sec:data}

We conducted experiments on five commonly used machine translation datasets, including multimodal machine translation dataset Multi30k~\citep{elliott2016Multi30K} English-to-German, Global Voices~\citep{GVoice} English-to-German , and WMT' 16 (100k) English-to-German (Newstest2016 as the test set)\footnote{
 To ensure a focused evaluation of the retrieved visual information's effectiveness, we intentionally sought to minimize the impact of data size. Consequently, we opted to construct our training set by randomly sampling 100,000 sentence pairs from the total pool of 4.5 million sentence pairs. This sampling approach aligns our dataset size more closely with that of other datasets for a fairer assessment.
}
, Bible~\citep{Christodou2015_bible} English-to-German, and MultiUN~\citep{eisele2010_multiun} English-to-German. 
The statistics for each dataset are presented in Table~\ref{tab:stat_data}.

\begin{table}
    \centering
    \begin{tabular}{|c||c|c|c|}
    \hline
    dataset & \!\!\!training set\!\!\! & \!\!\!dev set\!\!\! & \!\!\!test set\!\!\! \\ \hline \hline 
    Multi30k & 29,000 & 1,014 & 1,000 \\ \hline
    Global Voices & 69,227 & 2,000 & 2,000 \\  \hline
    WMT'16 (100k) & 100,000 & 2,000 & 3,000 \\ \hline
    Bible & 56,734 & 1,953 & 1,821 \\  \hline
    MultiUN & 56,235 & 4,000 & 4,000 \\ \hline
    \end{tabular}
    \caption{Statistics of datasets}
    \label{tab:stat_data}
\end{table}

\begin{table*}[ht]
    \centering
    \begin{tabular}{|l|l||c|}
    \hline
    \multicolumn{2}{|c||}{Method} & BLEU Score \\ \hline \hline 
    Text-only & Bi-LSTM~\citep{calixto2017doubly} & 33.70 \\ \cline{2-3}
    NMT & Transformer~\citep{zhang2019UVR} & 36.86 \\  \hline \hline
    & \citet{zhang2019UVR} & 36.86 \\ \cline{2-3}
    MMT with & \citet{zhao2021word} & 38.40 \\  \cline{2-3}
    Original Images & \citet{su2021bi-co} & 39.20 \\ \cline{2-3}
    & \citet{tang2022_multimodal} (Section~\ref{sec:tang_mmt}) & 38.14 \\ \hline \hline
    MMT with & \citet{zhang2019UVR} & 36.94 \\ \cline{2-3}
    Retrieved Images & \citet{tang2022_multimodal} (Section~\ref{sec:tang_mmt}) & 38.43 \\ \cline{2-3}
    & MMT with Visual Noise Filtering (Section~\ref{sec:filter_method}) & 38.51 \\ 
    \hline \hline
    \multicolumn{2}{|l||}{NMT with Retrieved Supplementary Text (Section~\ref{sec:supp_text_enc})} & \textbf{39.13} \\ 
    \hline
    \end{tabular}
    \caption{Results on Multi30K}
    \label{tab:bleu_Multi30K}
\end{table*}

\begin{table*}[ht]
    \centering
    \begin{tabular}{|l||c|c|c|c|c|}
    \hline
    Method & \multicolumn{5}{|c|}{Dataset} \\ \cline{2-6}
     & Multi30k & Global Voices & WMT`16 (100k) & Bible 
      & MultiUN \\ \hline \hline
    Text-only NMT & 33.70 & 9.22 & 7.99  & 35.23 & 39.49 \\ \hline
    MMT with Random Images & 37.65 & 9.29 & 8.11 & 35.31 & 39.48 \\ \hline
    MMT with Blank Images & 37.79 & 9.46 & 8.31 & 35.39 & 39.52 \\ \hline
    MMT with Retrieved Images & \textbf{38.43} & \textbf{9.81} & \textbf{8.41} & {\bf 35.42} & \textbf{39.53} \\ \hline
    \end{tabular}
    \caption{Translation performance across diverse datasets under varied image conditions (BLEU score)}
    \label{tab:image_compare}
\end{table*}

\subsection{Model Implementation}
\label{sec:model_impl}

For image retrieval, we used the Microsoft Bing\footnote{
  \url{https://global.bing.com/images}
} as the image search engine. 
In contrast, for supplementary text retrieval, we gathered sample sentences that included all the terms found in the respective search queries by referencing the Microsoft Bing Dictionary\footnote{
  \url{https://www.bing.com/dict}
}.
As described in Section~\ref{sec:img_ret} and Section~\ref{sec:text_ret}, we set $M$ to 5.
This choice signifies that we formulated 5 search queries and procured 5 images or supplementary text instances\footnote{
  When an insufficient number of sample sentences can be collected, we resort to large pretrained models like ChatGPT to generate sentences that meet the search query.
} for every source language sentence.

Regarding the text encoder, we used a bi-directional RNN with GRU to extract text features. Specifically, we used a 256 dimensional single-layer forward RNN and a 256 dimensional single-layer backward RNN.
For the translation decoder, we adhered to the approach proposed by ~\citet{su2021bi-co} and utilized a modified cGRU with hidden states of 256 dimensions.
Furthermore, we configured the embedding sizes for both source and target words to be 128.

As described in Section ~\ref{sec:img_enc}, the visual encoder we employed leveraged the $\mathrm{res4f}$ layer of a pretrained ResNet-50\citep{he2016res50} model to extract visual features of dimensions $196 \times 1024$.
Furthermore, as described in Section ~\ref{sec:supp_text_enc}, the supplementary text encoder utilized a BERT model pretrained on the BooksCorpus\citep{zhu2015aligning} and English Wikipedia\footnote{
  \url{https://en.wikipedia.org/wiki/English_Wikipedia}
}. 
This model was employed to extract textual features of dimensions $N \times 1024$, where $N$ represents the length of the retrieved supplementary text. 

Regarding the noise image filter, we set $M^{'}=10$ and used a CLIP model~\citep{radford2021learning} pretrained on the YFCC100M dataset~\citep{thomee2016yfcc100m} to filter out noisy images.
For the noise region fitler, we configured it with $O=128$. Here, we utilized a pretrained Faster R-CNN model~\citep{ren2015faster} that had been trained on the Open Images dataset~\citep{kuznetsova2020open}. This model was employed to identify and filter noisy regions in images effectively.

\subsection{Training Parameters}
\label{sec:train_para}

We initiated the word embeddings and other associated model parameters randomly, following a uniform distribution with a range of $-0.1$ to $0.1$.
During training, we employed the Adam optimizer with a mini-batch size of 32 and set the learning rate to 0.001. Additionally, a dropout strategy with a rate of 0.3 was applied to further enhance the models.
The training process continued for up to 15 epochs, with early stopping activated if the BLEU~\citep{papineni2002bleu} score on the development set did not exhibit improvement for 3 consecutive epochs. The model with the highest BLEU score on the dev set was selected for evaluation on the test set.
To minimize the impact of random seeds on experimental results and ensure result stability, we conducted the experiment 5 times with fixed random seeds and reported the macro-average of BLEU scores as the final result.

\subsection{Baselines}
\label{sec:baseline}

In the case of the Multi30k dataset, we conducted a quantitative comparison of the probing methods with several recent MMT models~\citep{zhang2019UVR,zhao2021word,su2021bi-co,tang2022_multimodal}.
However, the main focus of this research is to evaluate the necessity of visual information within real-world translation scenarios.
Four out of the five datasets utilized in our evaluation experiments are authentic text-only translation datasets without any visual annotation.
Consequently, for each dataset, we exclusively employed the text-only Bi-LSTM~\citep{calixto2017doubly} as a baseline.

The baseline model and the models detailed in Section~\ref{sec:models} were all trained using the same training set and identical training parameters. 
For all these models, we present the 4-gram BLEU score~\citep{papineni2002bleu} as the primary evaluation metric.

\section{Results and Analysis}
\label{sec:result}

Table~\ref{tab:bleu_Multi30K} presents the experimental results of the Multi30k dataset.  
Compared to various baseline models, all three probing methods mentioned in Section~\ref{sec:models} have achieved promising results. Notably, the MMT model with visual noise filtering (Section~\ref{sec:filter_method}) achieved a BLEU score of 38.51, while the NMT model with retrieved supplementary text (Section~\ref{sec:supp_text_enc}) achieved an impressive BLEU score of 39.13. 
In comparison to text-only NMT models~\citep{calixto2017doubly, vaswani2017transformer}, the NMT model with retrieved supplementary text significantly outperforms them, showcasing a substantial increase in BLEU score.
When compared to existing MMT methods that utilize original images~\citep{zhang2019UVR,zhao2021word,su2021bi-co}, the NMT model with retrieved supplementary text obtains a comparable BLEU score. 
Furthermore, in contrast to the MMT methods with retrieved images~\citep{zhang2019UVR,tang2022_multimodal}, the NMT model with retrieved supplementary text demonstrates performance gains of approximately 2.2 and 0.7 BLEU points, respectively.

Further experimental results and analysis will be presented in the following sections.

\begin{table*}[ht]
    \centering
    \begin{tabular}{|p{3.5cm}||c|c|c|c|c|}
    \hline 
     & Multi30k & Global Voices & WMT'16 (100k) & Bible & MultiUN \\ \hline \hline
     Number of sentences with half or more non-entity keywords & 27 & 94 & 796 & 398 & 818 \\ \hline
     Number of sentences with half of more noise images & 61 & 228 & 685 & 761 & 663  \\  \hline
    \end{tabular}
    \caption{Summary of manual analysis of image retrieval outcomes for each dataset}
    \label{tab:noise}
\end{table*}

\subsection{Translation Performances across Varied Datasets}
\label{sec:trans_result}

We firstly quantitatively compared text-only NMT~\citep{calixto2017doubly} with MMT utilizing retrieved images (Section~\ref{sec:tang_mmt}) across five diverse datasets mentioned in Section~\ref{sec:data}. 
As demonstrated in Table~\ref{tab:image_compare}, MMT achieved significantly higher BLEU scores on Multi30k, higher BLEU scores on Global Voices and WMT'16 (100k), and slightly higher BLEU scores on Bible and MultiUN.
It is intriguing to note that the improvement in translation performance is substantial on Multi30k, with an increase of approximately 4.7, whereas the gain on MultiUN is relatively modest, at approximately 0.04.

We speculate that the variations in results among the aforementioned translation datasets, such as Multi30k and other datasets, may be attributed to the differing qualities of images collected through the search engine.
To evaluate the influence of the quality of collected images, we train the MMT model with randomly retrieved unrelated images, blank images, and retrieved images from image search engines, respectively. 

The evaluation results are shown in table~\ref{tab:image_compare}.
It is obvious that MMT models with retrieved images achieves the highest BLEU score on all Multi30k and other four datasets, demonstrating the effectiveness of visual information from retrieved images.
Compared with the model with random images and blank images, the performance gain of collected images is approximately 0.7 \& 0.6 BLEU score on Multi30k, and 0.5 \& 0.3 BLUE score on Global Voices.
However, on WMT'16 (100k), Bible, and MultiUN datasets, models with retrieved images achieve almost the same BLEU score as the model with blank images.

One of the possible reason is that sentences from those three datasets contains fewer entity words that can be represented by images, and therefore, the search engine based image retrieval method collects numbers of noise images.
Sentences from WMT'16 (100k), Bible, and MultiUN datasets describe abstract concepts and complex events, while sentences from  
Multi30k and Global Voices describe real objects and people, which is more reliable for image retrieval.
\footnote{
  Examples of retrieved images from various datasets are presented in Table~\ref{tab:image_exp}.
}

To validate the hypotheses, we manually analyzed the image retrieval outcomes of each dataset.
In detail, we initially conducted a random sampling of 1,000 sentences and employed the image retrieval methods outlined in Section~\ref{sec:img_ret} to gather keywords and images for each sentence.
Regarding the extracted keywords, we conducted manual assessments to identify whether each keyword qualifies as an entity word.
Regarding the collected images, we carried out manual evaluations to determine if an image could offer pertinent visual information for the search query, and those that could not.
Images in the latter category were categorized as noise images.
Lastly, we tallied the quantity of sentences containing at least half of non-entity keywords and the quantity of sentences harboring at least half of noise images among the collected images.

As presented in Table~\ref{tab:noise}, for the Multi30k dataset, out of 1000 sentences, only 27 sentences contained half or more non-entity keywords,  and 61 sentences gathered half or more noise images from search engines. 
However, in the WMT'16 (100k) dataset, there are 796 sentences with half or more non-entity keywords and 685 sentences with half or more noise images, accounting for more than half of the sampled sentences. Consequently, our method shows poor performance on the WMT'16 (100k) dataset. The Bible dataset and MultiUN dataset exhibit a similar situation.
For the Global Voices dataset, there are 94 sentences with half or more non-entity keywords and 228 sentences with half or more noise images. This falls between the Multi30K and WMT'16 (100k) datasets. 
%
It is interesting to note that the Multi30k dataset, which has the smallest proportions of non-entity keywords and noise images, achieves the most significant gain in translation performance. On the other hand, datasets with the largest proportions of non-entity keywords and noise images show the smallest gain in translation performance.

\begin{table*}[ht]
    \centering
    \begin{tabular}{|l||c|c|c|c|c|}
    \hline 
     Method & \multicolumn{5}{|c|}{Dataset} \\ \cline{2-6}
      & Multi30k & Global Voices & WMT'16 (100k) & Bible & MultiUN \\ \hline \hline
     MMT with retrieved images & 38.43 & 9.81 & 8.41 & 35.42 & 39.53 \\ 
     \citep{tang2022_multimodal} &  &  &  &  & \\
     \hline \hline
     \ \ $+$ noise image filter & 38.50 & 10.12 & 8.89 & 36.12 & 39.91 \\ \hline
     \ \ $+$ noise region filter & 38.46 & 9.95 & 8.78 & 35.84 & 39.72 \\ \hline 
     \ \ $+$ noise image \& region filter & {\bf 38.51} & {\bf 10.23} & {\bf 8.93} & {\bf 36.38} & {\bf 39.95} \\ \hline
    \end{tabular}
    \caption{Results of image and region filtering method across diverse datasets (BLEU score)}
    \label{tab:noise_result}
\end{table*}

\subsection{Influence of the Correlation between Text and Images}
\label{sec:qua_influ}

Table~\ref{tab:noise_result} shows the evaluation results of applying two filtering approaches described in Section~\ref{sec:filter_method} in MMT.
It is obvious that MMT models with both noise image filter and noise region filter achieves the highest BLEU score across all datasets, including Multi30k and the other four,underscoring the effectiveness of these two filtering approaches.
\footnote{
  A correct example generated by MMT with visual noise filtering is presented in Table~\ref{tab:filter_exp}.
}

Notably, it is intriguing to note that the noise filtering techniques exhibited more substantial enhancements in translation performance for the WMT'16 (100k), Bible, and MultiUN datasets, in contrast to the improvements observed in the Multi30k and Global Voices datasets.
This further underscores the significant impact of the correspondence between image and text content on the translation performance the alignment and coherence between image and text content on the translation performance of the MMT system. It also elucidates why noise filtering methods yield marginal improvements on the Multi30K dataset.

In conclusion, the translation performance of the multimodal model primarily hinges on the consistency between textual and visual content. In other words, the more alignment exists between textual and visual content, the greater enhancement in translation performance with multimodal translation compared to text-only translation.
Hence, we arrive at a conclusion that aligns closely with \cite{caglayan2019probing}, which suggest that multimodal translation models predominantly treat visual information as a complement to textual information. 

\subsection{Exploring the Necessity of Visual Modality}
\label{sec:eff_vis_info}

\begin{table}
    \centering
    \begin{tabular}{|l||c|}
    \hline 
     Method & \!\!\!BLEU score\!\!\! \\ \hline 
     text-only NMT & 33.70 \\ \hline \hline
     \ \ $+$ visual information & 38.43  \\
     \ \ \ \ \ \ (MMT with retrieved images) & \\
     \ \ \ \ \ \ \citep{tang2022_multimodal} & \\ \hline 
     \ \ $+$ textual information & \textbf{39.13} \\ \hline
     \ \ $+$ visual \& textual information & 38.55 \\ \hline
    \end{tabular}
    \caption{Results on Multi30k using visual information or textual information enhanced NMT}
    \label{tab:text_sup_result}
\end{table}

We conducted a quantitative comparison between MMT with retrieved images (Section~\ref{sec:tang_mmt}) and NMT with retrieved supplementary texts on the Multi30k dataset.
Table~\ref{tab:text_sup_result} shows the experimental results. In comparison to MMT model employing images for translation enhancement, the approach integrating supplementary textual data for translation enhancement demonstrated a significantly higher BLEU score of 39.13. Remarkably, the combined utilization of both images and supplementary texts for translation enhancement yielded a BLEU score of 38.55, positioning itself between image-enhanced NMT and text-enhanced NMT. 

This demonstrates that both additional visual and  supplementary textual information play an entirely equivalent supplementary role in the translation process. Moreover, in most cases, the utilization of supplementary textual information assists the translation process more effectively.
\footnote{
  A correct example comparing NMT with retrieved supplementary texts to MMT with retrieved images is presented in Table~\ref{tab:text_exp}.
}

Therefore, we speculate that multimodal translation models trained on a large volume of data might face challenges in outperforming text-only translation models trained on comparable data volumes. This is because as the volume of data used in multimodal model training increases, the potential impact of visual information could diminish. We will verify this in future work.

\section{Conclusions}
\label{sec:col}
In this paper,  
we conduct an in-depth exploration into the role of visual information within the multimodal translation process on Multi30k and other four authentic translation datasets.
Our findings emphasize that the substantial correlation between visual and textual content significantly impacts the efficacy of multimodal translation, while employing filtering mechanisms based on the textual-visual correlation can enhance translation performance. 
Additionally, experimental results reveal that visual information plays a supplementary role in the multimodal translation process. This supplementary function of visual information can be substituted by the incorporation of supplementary textual information.
As one of our future work, we plan to assess the impact of the visual modality on more extensive translation datasets, including the complete WMT'16 dataset.
We speculate that as multimodal translation models are trained using larger datasets, the impact of visual information is likely to diminish.

\section*{Acknowledgements}
This research is supported by the Research Promotion Project of Key Construction Discipline in Guangdong Province (2022ZDJS112).

\nocite{*}
\section*{References}
\label{sec:reference}
\bibliography{custom}
\bibliographystyle{lrec-coling2024-natbib} 

\appendix
\section{Qualitative Examples}
\label{sec:app_exp_img}

In this appendix, we provide examples of retrieved images  (Table~\ref{tab:image_exp}), as well as translation examples for MMT with visual noise filtering (Table~\ref{tab:filter_exp}) and NMT with retrieved supplementary texts (Table ~\ref{tab:text_exp}).

\begin{table*}
    \centering
    \begin{tabular}{|c|l|c|}
        \hline
        Dataset & English Sentence & One of five retrieved images \\ \hline \hline
        Multi30k 
          &  \begin{tabular}{p{4.5cm}}
            The person in the striped shirt is mountain climbing. \\
          \end{tabular}
          & \begin{tabular}{c}
            \specialrule{0em}{1pt}{1pt}
            \includegraphics[scale=0.22]{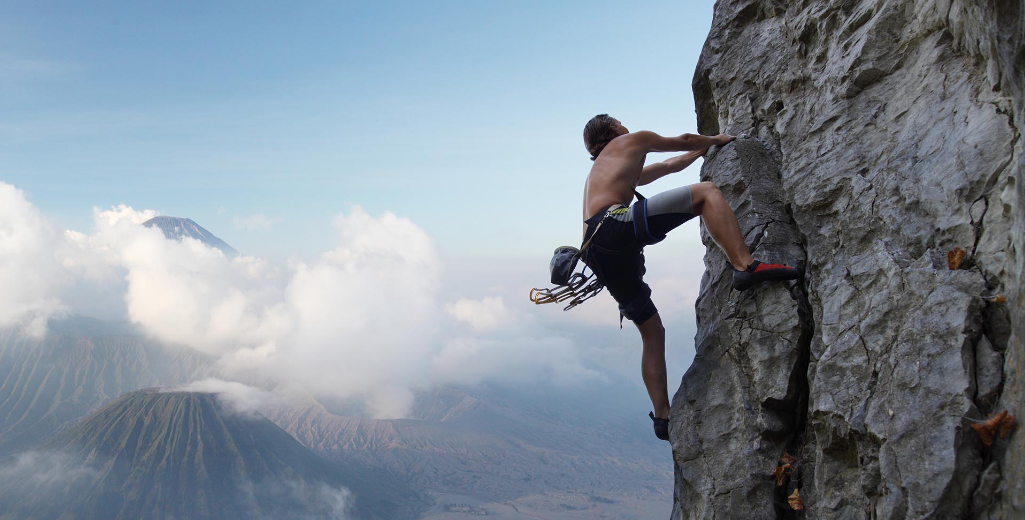}
            \end{tabular} \\ \hline
         \rule[-2pt]{0mm}{1.9cm}
        Global Voices 
          & \begin{tabular}{p{4.5cm}}
            Now the city is under a siege from the security forces. \\
          \end{tabular} 
          & \begin{tabular}{c}
            \includegraphics[scale=0.35]{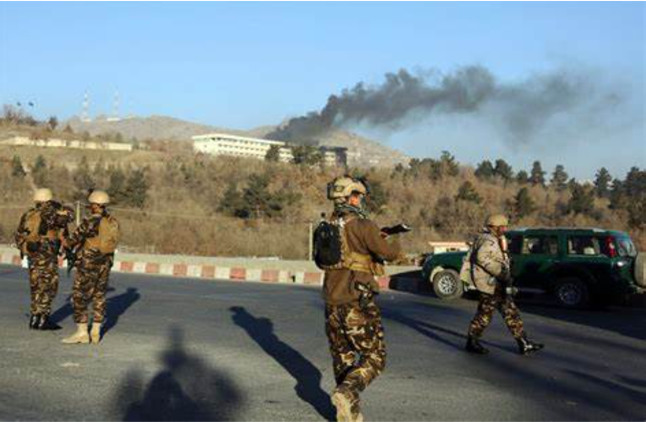}
           \end{tabular} \\ \hline
        \rule[-2pt]{0mm}{1.9cm}
        WMT'16 (100k) 
        & \begin{tabular}{p{4.5cm}}
            In the future, integration will be a topic for the whole of society even more than it is today. \\
          \end{tabular}
        &  \begin{tabular}{c}
           \includegraphics[scale=0.35]{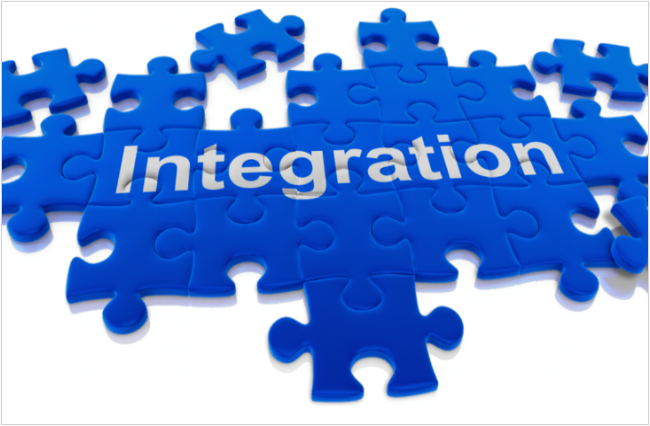}
          \end{tabular} \\ \hline
        Bible 
        & \begin{tabular}{p{4.5cm}}
             You are Yahweh, even you alone. You have made heaven. the heaven of heavens, with all their army, the earth and all things that are on it, the seas and all that is in them and you preserve them all. \\
          \end{tabular}
        & \begin{tabular}{c}
            \includegraphics[scale=0.22]{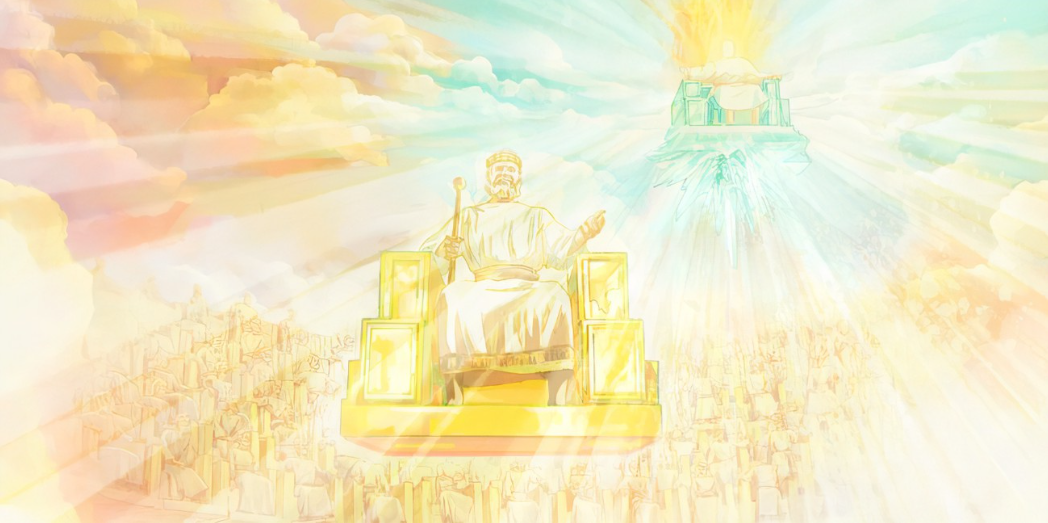}
          \end{tabular} \\ \hline
        MultiUN 
        & \begin{tabular}{p{4.5cm}}
          Development assistance cannot by itself prevent or end conflict. \\
        \end{tabular} 
        & \begin{tabular}{c}
          \includegraphics[scale=0.13]{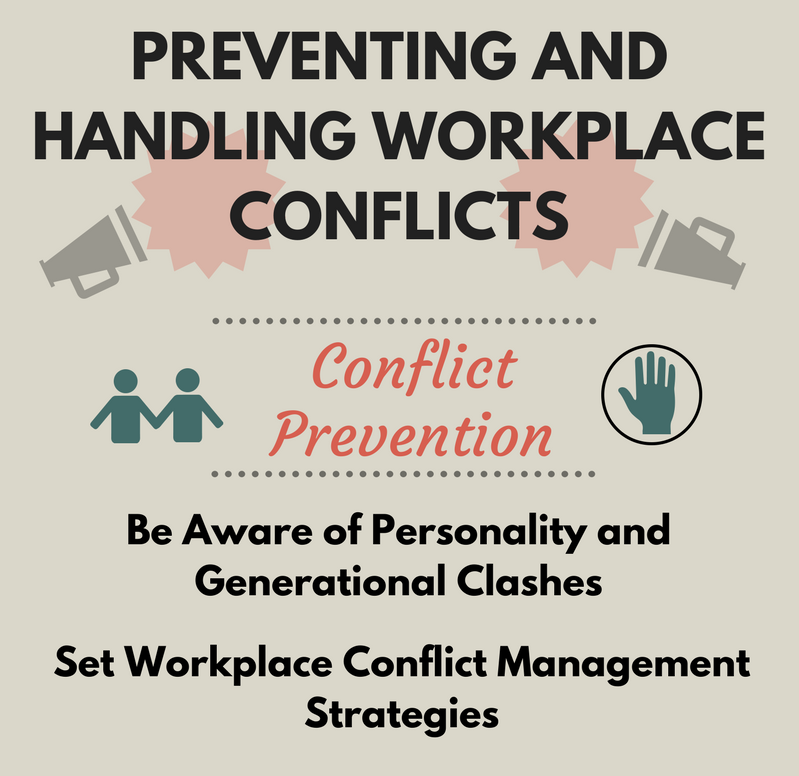}
        \end{tabular} \\ \hline
    \end{tabular}
    \caption{Examples of retrieved image from different datasets. For the sentence from Multi30k dataset, our method efficiently retrieves an image that accurately represents the sentence's content ``A man is rock climbing''. For the sentence from Global Voice dataset, the retrieved image exhibits a degree of alignment with the source sentences, encompassing elements like ``city'',``siege'' and``forces''. However, for the sentence from WMT'16 (100k), Bible and MultiUN datasets, it becomes evident that the retrieved images offer limited relevant visual information and thus provide little assistance for translation.}
    \label{tab:image_exp}
\end{table*}

\begin{table*}
\centering
    \begin{tabular}{|p{4cm}||p{10cm}|}
      \hline
      Source (En) & But he answered and said, "Every plant which my heavenly Father didn't plant will be uprooted. \\ \hline
      Target (De) & Aber er antwortete und sprach: Alle Pflanzen, die mein himmlischer Vater nicht pflanzte, die werden ausgereutet. \\ \hline \hline
      Retrieved images & 
       \begin{tabular}{ccc}
          \includegraphics[scale=0.12]{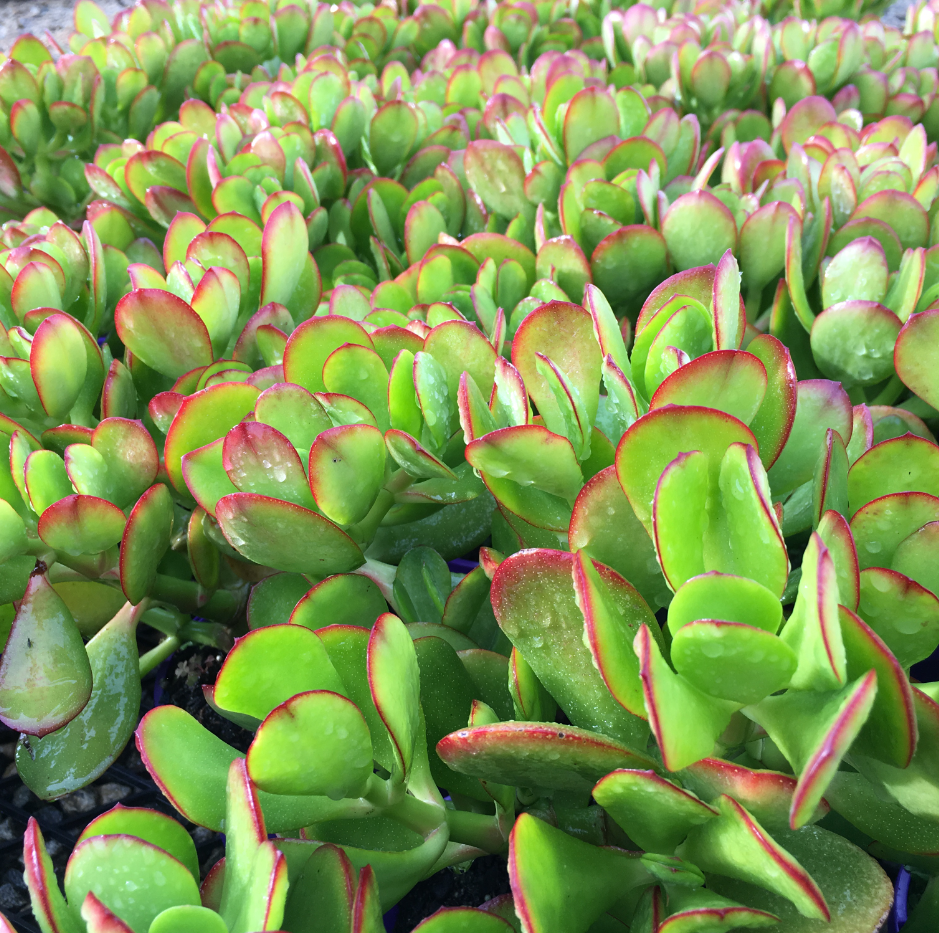}
         & \includegraphics[scale=0.21]{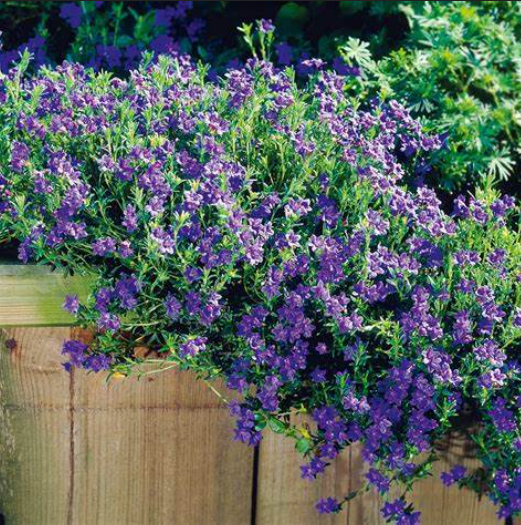}
         & \includegraphics[scale=0.25]{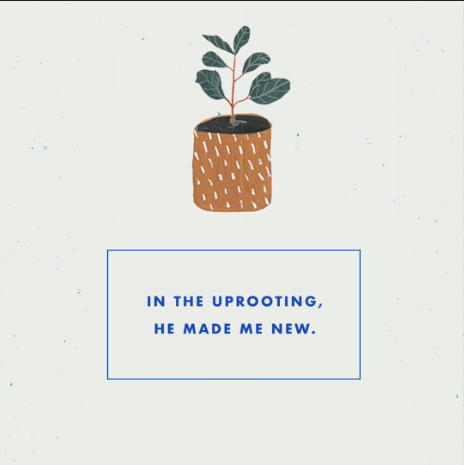} \\
         \includegraphics[scale=0.3]{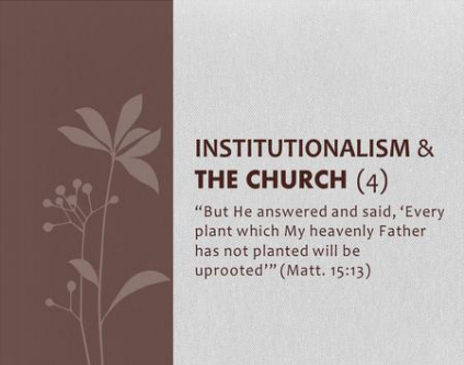}
         & \includegraphics[scale=0.22]{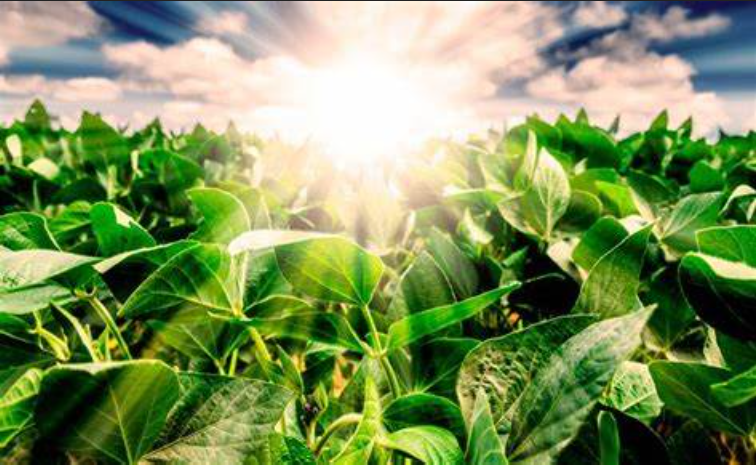} & 
        \end{tabular} \\ \hline
      MMT with retrieved images & Er antwortete aber und sprach: Alle Pflanzen, die mein himmlischer Vater nicht verderbte Quelle. \\ \hline \hline
      Retrieved images with noise image filter & 
      \begin{tabular}{ccc}
          \includegraphics[scale=0.22]{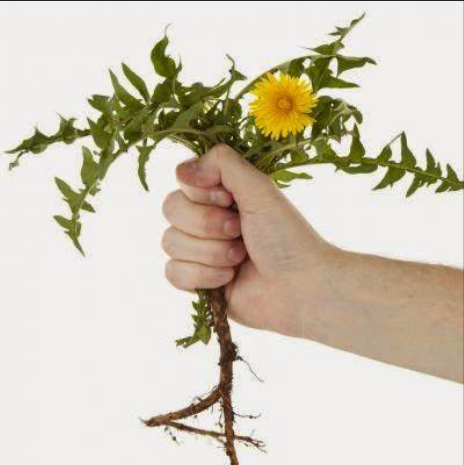}
         & \includegraphics[scale=0.22]{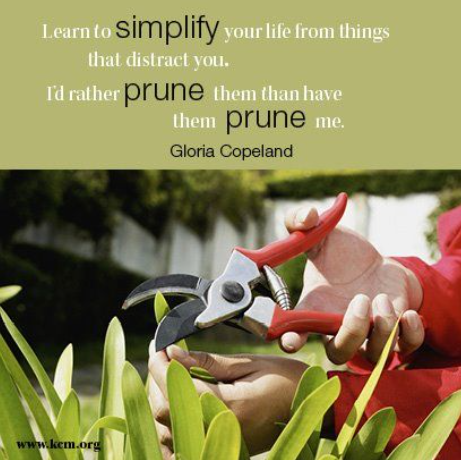}
         & \includegraphics[scale=0.22]{figures/prefilter_5.png} \\
         \includegraphics[scale=0.12]{figures/prefilter_1.png}
         & \includegraphics[scale=0.21]{figures/prefilter_2.png} & 
      \end{tabular} \\ \hline
      MMT with noise image filter & Er antwortete aber und sprach: Alle Pflanzen, die mein himmlischer Vater nicht pflanzte. \\ \hline
      MMT with both noise image and region filter & Er antwortete aber und sprach: Alle Pflanzen, die mein himmlischer Vater nicht pflanzte,  wird entwurzelt werden. \\ \hline
    \end{tabular}
    \caption{A correct example generated by MMT with visual noise filtering. Due to its unique characteristics, the Bible dataset contains numerous entity words but is challenging to obtain images that effectively represent the textual content. However, visual noise filtering based on visual-text correlation can partially alleviate this situation. In this example, the filtered visual information has enabled the translation of ``uprooted'' to be correct.}
    \label{tab:filter_exp}
\end{table*}

\begin{table*}
\centering
    \begin{tabular}{|p{4cm}||p{10cm}|}
      \hline
      Source (En) & Group of Asian boys wait for meat to cook over barbecue. \\ \hline
      Target (De) & Eine Gruppe asiatischer Jungen wartet am Grill darauf, dass Fleisch gar wird. \\ \hline \hline
      Text-only NMT & Eine asiatische Jungen warten auf dem Fleisch, um den Grill zu kochen. \\ \hline \hline
      Retrieved images & 
        \begin{tabular}{ccc}
          \includegraphics[scale=0.1]{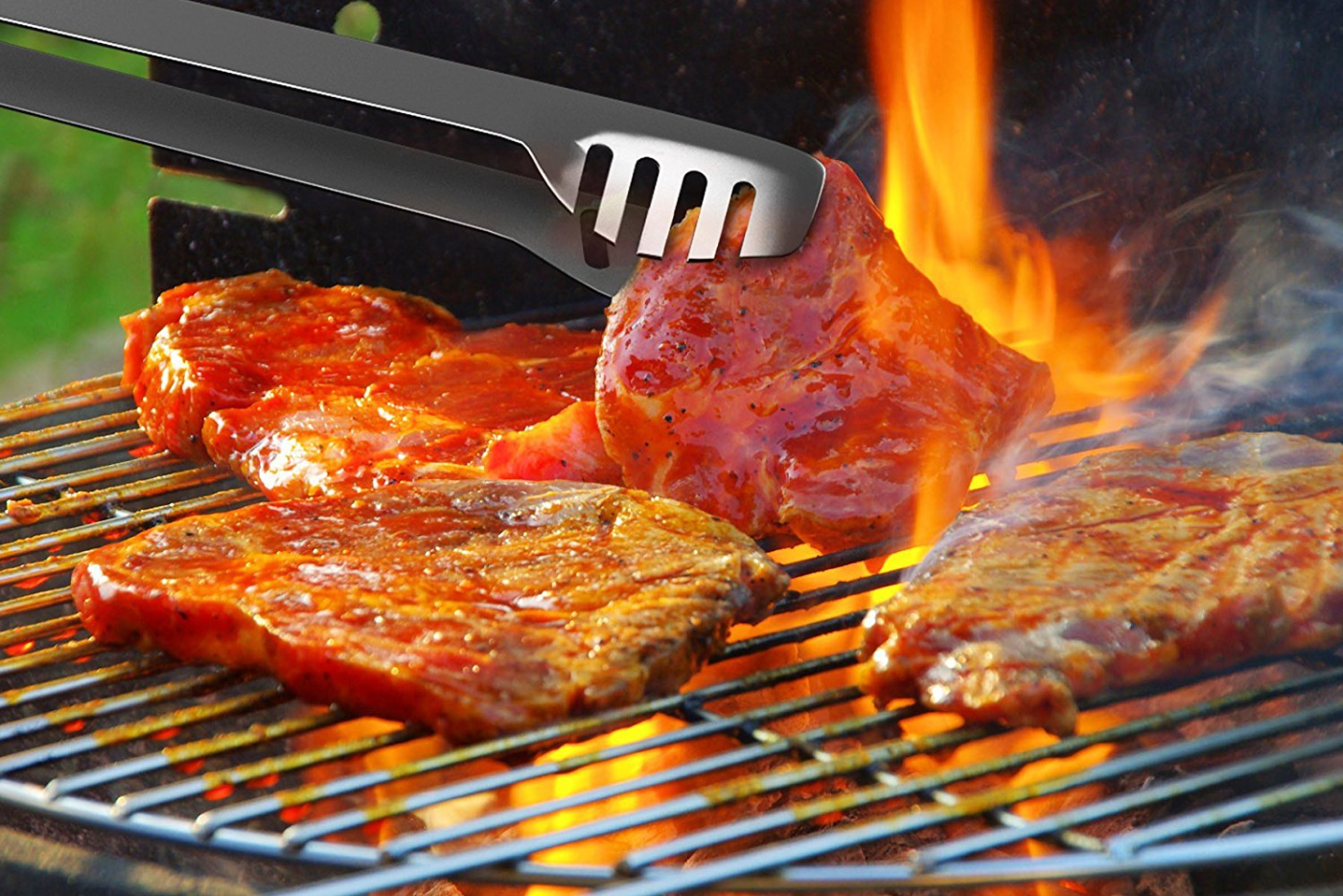}
         & \includegraphics[scale=0.11]{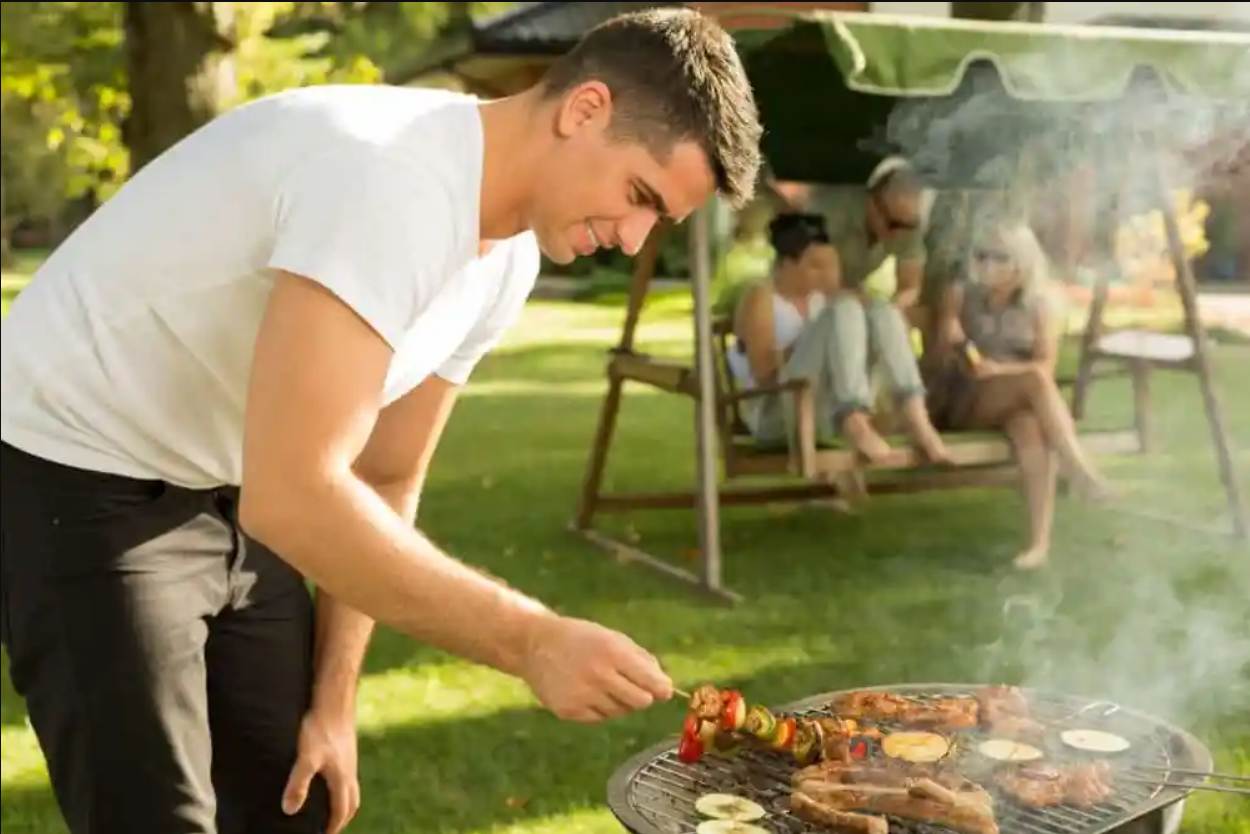}
         & \includegraphics[scale=0.14]{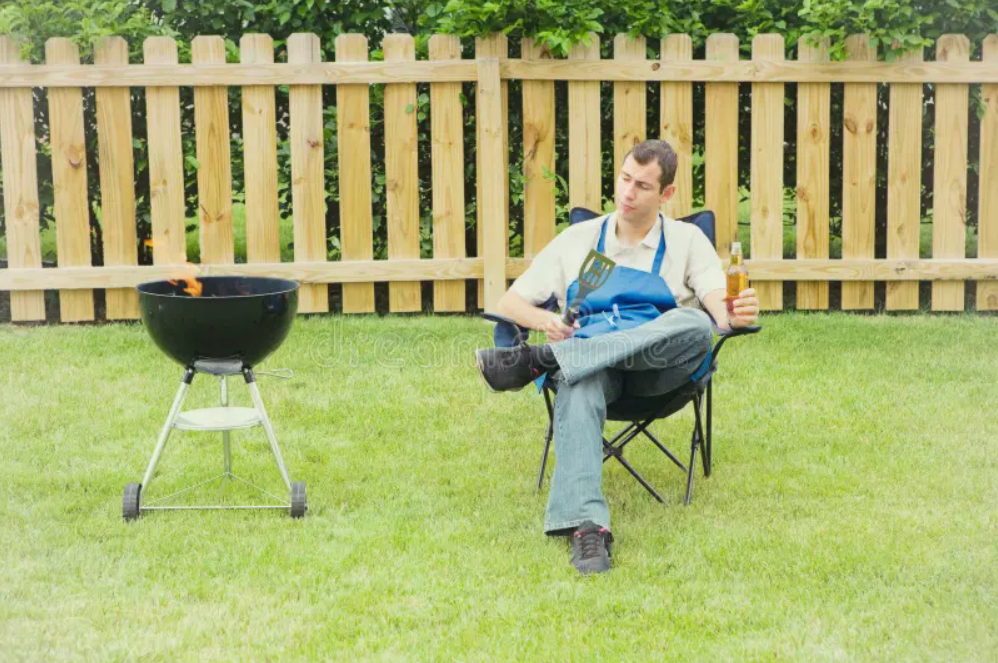} \\
         \includegraphics[scale=0.14]{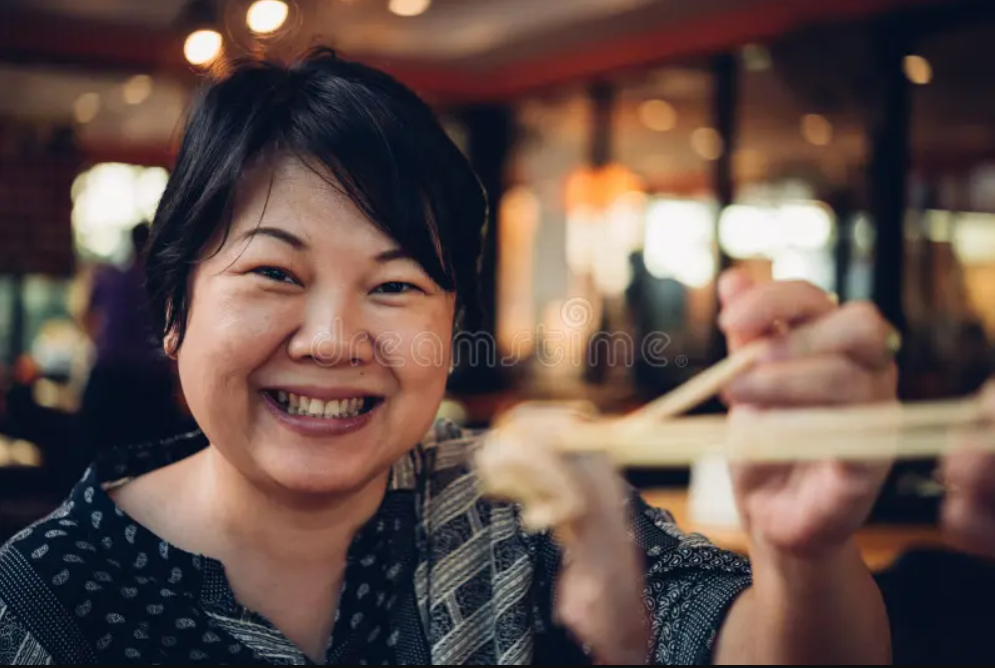}
         & \includegraphics[scale=0.16]{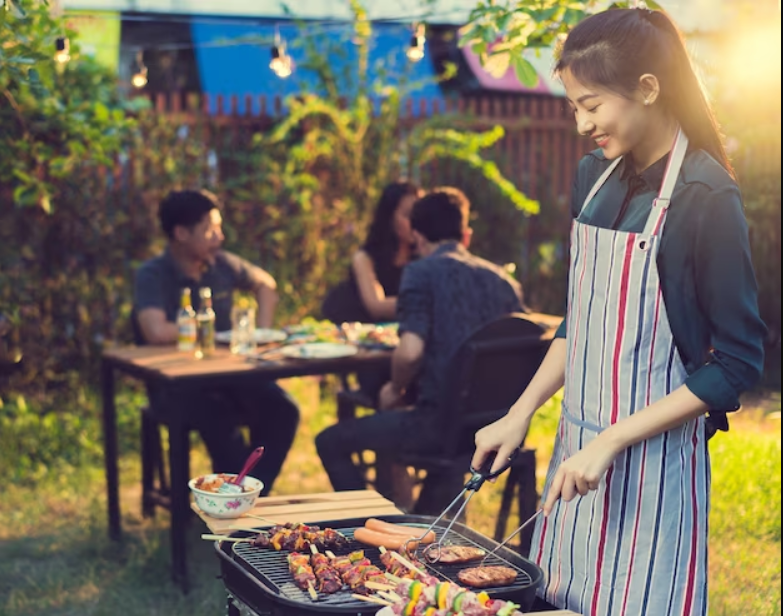} & 
        \end{tabular} \\ \hline
      MMT with retrieved images & Eine Gruppe von asiatischen Jungen wartet darauf, um Fleisch zu grillen. \\ \hline \hline
      Retrieved supplementary texts & 
       \begin{tabular}{rp{9.3cm}}
         \!\!\!(1)\!\!\! & Delivery is hardly limited to pizza at this point; everything from sushi to barbecue seems available as a to-go order. \\
         \!\!\!(2)\!\!\! & While the savory aroma of barbecue filled the air, friends and family gathered around the grill, eagerly sharing stories and laughter as they waited for the delicious meal to be ready. \\
         \!\!\!(3)\!\!\!& As the sun dipped below the horizon, our group of friends decided to have a barbecue in the backyard, lighting up the grill and eagerly waiting for the charcoal to heat up so that we could start cooking our favorite dishes. \\
         \!\!\!(4)\!\!\! & At the lively outdoor barbecue gathering, a diverse group of friends, including a talented Asian chef, couldn't wait to cook up a mouthwatering feast. \\
         \!\!\!(5)\!\!\! & While the enthusiastic Asian group gathered around the barbecue, they took turns to cook their favorite dishes, making everyone else eagerly wait in anticipation of the delicious meal. \\
       \end{tabular}
      \\ \hline
      MMT with retrieved supplementary texts & Eine Gruppe von asiatischen Jungen wartet darauf, dass Fleisch \"uber Grill zukochen. \\ \hline
    \end{tabular}
    \caption{A correct example generated by NMT with retrieved supplementary texts. In this example, in contrast to text-only NMT without any supplementary information, visual information and supplementary text information play an equivalent role, correctly translating ``Group'' to ``Gruppe''. Benefiting from the rich information in the supplementary text, the NMT with retrieved supplementary text achieves more accurate translations compared to MMT with retrieved images.}
    \label{tab:text_exp}
\end{table*}

\end{document}